\documentclass[11pt,letterpaper]{article}
\usepackage[margin=1in]{geometry}
\usepackage{times}
\usepackage{parskip}
\usepackage{natbib}
\setcitestyle{authoryear,round,citesep={;},aysep={,},yysep={;}}

\usepackage{amsmath,amsfonts,bm}

\def\eqref#1{equation~\ref{#1}}

\def\1{\bm{1}}

\def\vk{{\bm{k}}}

\def\vp{{\bm{p}}}
\def\vq{{\bm{q}}}

\def\mR{{\bm{R}}}

\DeclareMathAlphabet{\mathsfit}{\encodingdefault}{\sfdefault}{m}{sl}
\SetMathAlphabet{\mathsfit}{bold}{\encodingdefault}{\sfdefault}{bx}{n}

\usepackage{url}
\usepackage{amsmath,amssymb}
\usepackage{booktabs,multirow}
\usepackage{graphicx}
\usepackage{algorithm}
\usepackage{algpseudocode}
\usepackage{microtype}
\usepackage{enumitem}
\usepackage{float}
\usepackage[pdfusetitle]{hyperref}
\hypersetup{hidelinks}
\newcommand{\BenchCudaQwen}{12.1}
\newcommand{\BenchFlashAttnQwen}{2.7.4.post1}
\newcommand{\BenchNativeCommit}{5d84729}
\newcommand{\BenchTorchQwen}{2.3.0}
\newcommand{\BenchTransformersQwen}{4.51.3}
\newcommand{\BenchTritonQwen}{3.4.0}
\newcommand{\LbdKvzipLlamaSevenEighths}{92.8}
\newcommand{\LbdKvzipQwenSevenEighths}{92.2}
\newcommand{\RuLlamaEchoSevenEighths}{92.9}
\newcommand{\RuQwenEchoSevenEighths}{92.1}

\newcommand{\FcQueryFirstQwen}{0.154}
\newcommand{\FcQueryLaterQwen}{0.141}
\newcommand{\FcRhoFirstQwen}{0.678}
\newcommand{\FcRhoLaterQwen}{0.692}
\newcommand{\FcMedianRhoFirstQwen}{0.717}
\newcommand{\FcMedianRhoLaterQwen}{0.718}
\newcommand{\FcPrefixZeroQwen}{0.158}
\newcommand{\FcPrefixFullQwen}{0.143}

\newcommand{\FcQueryFirstLlama}{0.114}
\newcommand{\FcQueryLaterLlama}{0.099}
\newcommand{\FcRhoFirstLlama}{0.735}
\newcommand{\FcRhoLaterLlama}{0.745}
\newcommand{\FcMedianRhoFirstLlama}{0.772}
\newcommand{\FcMedianRhoLaterLlama}{0.770}
\newcommand{\FcPrefixZeroLlama}{0.117}
\newcommand{\FcPrefixFullLlama}{0.105}

\renewcommand{\eqref}[1]{(\ref{#1})}
\newcommand{\method}{EchoPress}
\newcommand{\kvzip}{KVzip}

\DeclareRobustCommand{\stepnum}[1]{\textcircled{\scriptsize #1}}

\title{\method{}: Query-Agnostic KV Cache Pruning\\via Virtual Context Reconstruction}

\author{
Jiawei Lin \qquad Saibo Geng \qquad Thomas Bourgeat \\
EPFL \\
\texttt{\{jiawei.lin,saibo.geng,thomas.bourgeat\}@epfl.ch}
}
\date{}

\begin{document}

\maketitle

\begin{abstract}
KV cache pruning reduces long-context inference memory usage by evicting less
important key--value pairs. \kvzip{} estimates importance through context
reconstruction: prompting a model to repeat the context chunk by chunk.
This achieves strong compression quality at the cost of additional forward
passes. Learned approximations reduce this
cost but require model-specific training.
We analyze how \kvzip{} identifies important cached information and show how
to approximate its reconstruction scores using information already computed
during prefill. These findings motivate \method{}, a training-free method that
approximates reconstruction attention using queries and keys from standard
prefill. For each request, it reconstructs only the first chunk to calibrate
importance scores for the remaining context.
Experiments on LongBench and RULER with Qwen3-8B and Llama-3.1-8B-Instruct
show that \method{} matches \kvzip{} in task accuracy across eviction ratios
from 50\% to 90\%, while reducing compression overhead by 1.7--19.6$\times$ and
total prefill time by up to 2.9$\times$.
Code is available at
\href{https://github.com/ljwljwljwljw/kvpress/tree/echo-press}{\texttt{https://github.com/ljwljwljwljw/kvpress}}.

\end{abstract}

\begin{figure}[tbp]
\centering
\includegraphics[width=0.9\linewidth]{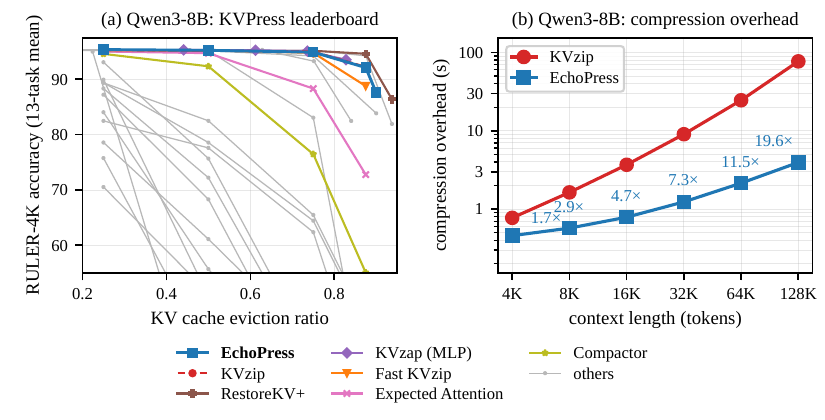}
\caption{
    (a) Qwen3-8B RULER-4K accuracy (13-task mean) for query-agnostic
    methods from the KVPress leaderboard \citep{hsieh2024ruler,kvpress2024},
    with our \method{} results overlaid. RestoreKV(+), KVzap, and Fast KVzip
    use model-specific training
    \citep{baek2026restorekv,jegou2026kvzap,kim2026fastkvzip}.
    \method{} requires neither training nor external calibration data.
    (b) Qwen3-8B compression overhead (additional time beyond plain prefill,
    including cache compaction) on one 80\,GB A100 GPU at 50\% eviction. The \kvzip{}
    baseline uses its official CUDA
    implementation \citep{kim2025kvzip}.
}
\label{fig:teaser}
\end{figure}

\section{Introduction}
\label{sec:intro}

The key--value (KV) cache grows linearly with sequence length, increasing
memory use during long-context inference \citep{kwon2023pagedattention}.
KV eviction reduces this cost by discarding
less important KV pairs. \emph{Query-aware} methods use attention from a
question or recent observation window to choose what to retain
\citep{li2024snapkv,cai2024pyramidkv}, but one compressed cache may not serve a
different question reliably \citep{li2025scbench}. \emph{Query-agnostic}
compression uses only the context, allowing the result to be reused across
questions, conversation turns, and shared prefixes
\citep{gim2024promptcache,zheng2024sglang}.

Existing query-agnostic methods trade scoring quality for cost. Prefill-based
methods use key norms \citep{devoto2024knorm}, key geometry
\citep{park2025keydiff,godey2025qfilters}, or expected attention
\citep{devoto2025expectedattention} without another model forward pass. Reconstruction
methods such as \kvzip{} \citep{kim2025kvzip} instead prompt the model to repeat
the context chunk by chunk after prefill.
Each KV pair's importance is scored by the maximum attention it receives during
this repetition.
A global threshold then evicts the least important pairs across all layers and
KV heads.
Reconstruction achieves state-of-the-art compression quality but requires
expensive additional forward passes over the entire context.
In our 4K--128K experiments, \kvzip{}'s compression overhead is
2.1--2.2$\times$ the matching plain-prefill time
(Section~\ref{sec:latency}).
Training-based methods replace reconstruction with model-specific
score predictors or eviction gates \citep{jegou2026kvzap,kim2026fastkvzip},
reducing inference cost but requiring a separate training stage.

We ask whether prefill states can approximate reconstruction scores without
training a predictor. Reconstruction differs from prefill in three respects:
(1) each repetition query can attend to every cached key in its original chunk,
including keys that followed it in the original token order, while attention
within the repeated copy remains causal;
(2) repetition queries are positioned after the cached context, shifting their
RoPE coordinates relative to prefill \citep{su2024roformer}; and
(3) reconstruction queries are formed from hidden states conditioned on the
full context rather than only a causal prefix.
Through these hidden states, context keys outside the target chunk can influence
reconstruction queries even when excluded from \kvzip{}'s scoring softmax.
Our experiments show that exposing all cached keys in
the original chunk (difference 1) modestly improves agreement between our
prefill-based scores and \kvzip{}'s reconstruction scores, while correcting the
positional shift (difference 2) gives the largest gain. Together,
these two corrections raise the mean per-request median Spearman correlation
from 0.208 to 0.678 on Qwen3-8B and from 0.237 to 0.740 on
Llama-3.1-8B-Instruct (Section~\ref{sec:analysis}).

\begin{figure}[t]
  \centering
  \includegraphics[width=\linewidth]{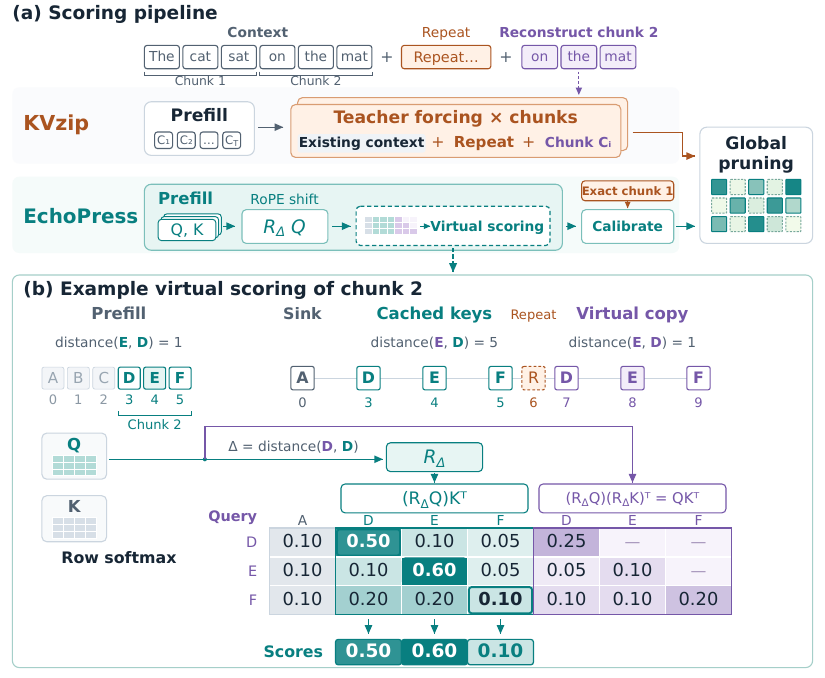}
  \caption{
  \method{} scoring pipeline and virtual reconstruction.
  (a) \kvzip{} reconstructs every chunk. \method{} reconstructs only the first,
  reuses prefill states to score later chunks, and calibrates their scores using
  the first chunk for each layer and KV head before global pruning (dashed cells).
  (b) Virtual scoring combines \stepnum{1} full visibility of teal cached keys
  in the target chunk, \stepnum{2} RoPE-shifted queries, and \stepnum{3} a causal
  purple virtual copy. Shared rotations cancel in the copy, allowing reuse of
  $QK^\top$. Prompt R determines the offset but is omitted from scoring.
  For each query, a joint softmax normalizes attention over sink, cached, and
  causally visible copy keys. Each cached key's raw score is the largest
  attention probability it receives across query positions and query heads
  sharing its KV head. The probabilities shown are illustrative; the full
  attention matrix and repeated cache are never materialized.}
  \label{fig:overview}
\end{figure}

These findings motivate \method{}'s \emph{virtual context reconstruction}
(Figure~\ref{fig:overview}). It reuses prefill queries and cached keys, corrects
the queries' RoPE positions, and adds a causal virtual-copy branch.
Virtual scores preserve useful within-head rankings, but their scales differ
across layers and KV heads, complicating global eviction. First-chunk prefill
queries also differ most from reconstruction queries because they have seen the
least preceding context. \method{} therefore reconstructs the first chunk
exactly and pairs its exact and virtual scores to construct a per-layer,
per-KV-head quantile map. The map places later virtual scores on the
exact scale, so each request self-calibrates without training. To compute virtual
scores efficiently for all later chunks, we implement the scoring operation
with fused Triton kernels \citep{tillet2019triton}.

Our contributions are:
\begin{itemize}[leftmargin=*,itemsep=1pt,topsep=2pt,parsep=0pt]
\item An analysis of \kvzip{}'s reconstruction scoring that separates key
visibility, rotary positions, and context conditioning. It identifies the role
of position correction in recovering within-head rankings and of score-scale
calibration in global eviction.
\item \method{}, a training-free compression method derived from this analysis.
A single exact first-chunk reconstruction supplies request-specific calibration
for each layer and KV head, requiring no external calibration data or stored calibration
artifacts.
\item Evaluation on LongBench and RULER-4K with Qwen3-8B and
Llama-3.1-8B-Instruct. \method{} stays within 0.9 points of \kvzip{} at
50--75\% eviction, exceeds it on all four model--benchmark pairs at 90\%, and
reduces compression overhead by 1.7--19.6$\times$ on an A100 GPU at
4K--128K tokens.
\end{itemize}

\section{\method{}: virtual context reconstruction}
\label{sec:method}

\method{} approximates \kvzip{}'s reconstruction scores from queries and keys
computed during prefill. It scores every chunk virtually, reconstructs the
first chunk exactly, and uses the paired first-chunk scores to calibrate later
virtual scores. Figure~\ref{fig:overview} summarizes the method.

\subsection{Reconstruction scoring in \kvzip{}}
\label{sec:method:reference}

Consider a decoder with $L$ layers, $H$ query heads, and $H_{kv}$ KV heads of
dimension $d$, with $G=H/H_{kv}$ query heads sharing each KV head
\citep{ainslie2023gqa}. The prefilled
cache contains $n$ tokens, partitioned into chunks
$\mathcal C_t=[a_t,b_t)$ of length $m_t\le C$. For each chunk, \kvzip{}
\citep{kim2025kvzip} runs a teacher-forced forward pass using the full cache as
a prefix. Its input contains a repeat prompt $\vp_t$ of length $P_t$
(e.g., ``Repeat the part of the previous context exactly''), followed
by the original chunk. The temporary reconstruction cache is discarded after
each chunk. Although this forward pass uses the full original cache, \kvzip{}
forms its importance softmax from only the initial sink keys, the target cached
chunk, and the causal prefix of the repeated chunk. Other context keys affect
the reconstruction hidden states but do not enter the scoring softmax.

We use the scoring rule implemented in KVPress \citep{kvpress2024}. To make the
conditioning difference explicit, we reserve unadorned $\vq,\vk$ for vectors
produced during the original causal prefill and use $\tilde\vq,\tilde\vk$ for
vectors recomputed by \kvzip{} after the full original cache has been supplied
as a prefix. Thus, suppressing layer and KV-head indices,
$\tilde\vq_u^g,\tilde\vk_u$ are full-context-conditioned reconstruction
vectors at position $n+u$, whereas $\vk_i$ is the causally computed key stored
for original context position $i$. All queries and keys include their rotary
embeddings. The score for cached position $i\in\mathcal C_t$ is
\begin{align}
\label{eq:kvzip}
s_i^{\ell h}
&=\max_{\substack{1\le g\le G\\0\le u<P_t+m_t}}
  \frac{\exp(\tilde\vq_u^g\!\cdot\vk_i/\sqrt d)}{Z_u^g},\\
Z_u^g
&=\sum_{v\in\mathcal S_t\cup\mathcal C_t}
  \exp(\tilde\vq_u^g\!\cdot\vk_v/\sqrt d)
  +\sum_{v=0}^{u}\exp(\tilde\vq_u^g\!\cdot\tilde\vk_v/\sqrt d),\nonumber
\end{align}
where $\mathcal S_t=[0,\min(4,a_t))$ contains initial sink keys. At eviction
ratio $r$, a global threshold removes the $\lfloor rLH_{kv}n\rfloor$
lowest-scoring pairs across layers, heads, and positions.

\subsection{Virtual reconstruction from prefill states}
\label{sec:method:virtual}

At a high level, virtual reconstruction replaces a \kvzip{} reconstruction
pass with an attention calculation over states already produced during
prefill. For each virtually scored chunk, \method{} retains the causal prefill
representations, shifts them to the RoPE positions where the repeated tokens
would appear, and evaluates two key blocks jointly: the original cached chunk
and a causal virtual copy.

Let $\vq_j^g,\vk_j$ denote the unadorned prefill vectors at original context
position $j\in\mathcal C_t$. These vectors are conditioned only on the causal
prefix through $j$, unlike the full-context-conditioned
$\tilde\vq,\tilde\vk$ in Section~\ref{sec:method:reference}. They also use a
different RoPE position: reconstruction shifts token $j$ by
$\Delta_t=n+P_t-a_t$. \method{} retains the causal prefill states and reproduces
this positional shift exactly. We use a hat to denote the resulting virtual
vectors:
\begin{equation}
\label{eq:offset}
\hat\vq_j^g=\mR_{\Delta_t}\vq_j^g,
\qquad
\hat\vk_j=\mR_{\Delta_t}\vk_j.
\end{equation}
The hats therefore do not represent newly computed full-context states.
Instead, they are rotated versions of the original prefill vectors. A shared rotation
preserves dot products, giving
\begin{equation}
\label{eq:blocks}
\hat\vq_j^g\!\cdot\vk_i=(\mR_{\Delta_t}\vq_j^g)\!\cdot\vk_i,
\qquad
\hat\vq_j^g\!\cdot\hat\vk_i
=(\mR_{\Delta_t}\vq_j^g)\!\cdot(\mR_{\Delta_t}\vk_i)
=\vq_j^g\!\cdot\vk_i.
\end{equation}
Thus the cached chunk and its virtual copy share one key tensor. The first
branch uses shifted queries against cached keys, whereas the second reuses unshifted
query--key products for the virtual copy.

Dropping the repeat prompt's query rows and key columns yields
\begin{align}
\label{eq:virtual}
\hat s_i^{\ell h}
&=\max_{\substack{1\le g\le G\\j\in\mathcal C_t}}
  \frac{\exp(\hat\vq_j^g\!\cdot\vk_i/\sqrt d)}{\hat Z_j^g},\\
\hat Z_j^g
&=\sum_{v\in\mathcal S_t\cup\mathcal C_t}
  \exp(\hat\vq_j^g\!\cdot\vk_v/\sqrt d)
  +\sum_{\substack{v\in\mathcal C_t\\v\le j}}
  \exp(\vq_j^g\!\cdot\vk_v/\sqrt d).\nonumber
\end{align}
Each shifted query sees the whole cached chunk, while its virtual-copy branch
remains causal. The repeat prompt determines $P_t$ in \eqref{eq:offset} but
does not need to be processed by the model. Like \eqref{eq:kvzip},
\eqref{eq:virtual} is chunk-local. The remaining approximation error comes from
reusing causal prefill states and omitting the prompt from the scoring softmax.

\subsection{Per-request first-chunk calibration}
\label{sec:method:scale}

\begin{figure}[htbp]
\centering
\includegraphics[width=0.78\linewidth]{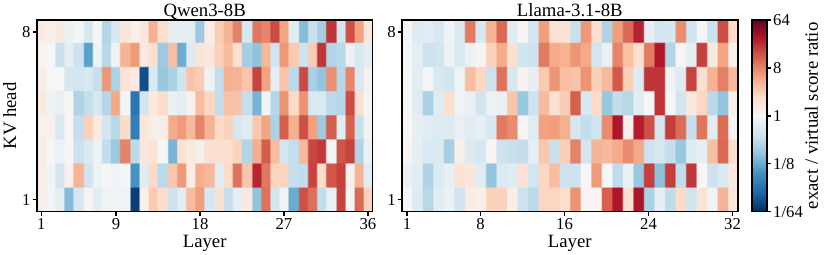}
\caption{Virtual-score scale distortion across layers and KV heads. For each
context and cell, we divide the exact first-chunk score quantile at $q=0.9$ by
the corresponding virtual-score quantile. Each heatmap shows the median ratio
over eight LongBench contexts.}
\label{fig:calibration_scale}
\end{figure}

Virtual scores are useful within a head, but their scale distortion varies
across layers and KV heads. Since global eviction compares all scores with one
threshold, this distortion can assign the wrong cache budget to a layer--KV-head
pair.
Figure~\ref{fig:calibration_scale} shows this variation: the ratio between
exact and virtual scores changes substantially across layers and KV heads.
\method{} therefore uses the request itself to align each layer--KV-head pair's
scale.

During prefill, we retain the virtual scores $\hat s_i^{\ell h}$ for the first
chunk. We then run one exact reconstruction pass for that chunk, producing
$s_i^{\ell h}$. For every layer $\ell$ and KV head $h$, let
$\hat s_{(1)}^{\ell h}\le\cdots\le\hat s_{(m_1)}^{\ell h}$ and
$s_{(1)}^{\ell h}\le\cdots\le s_{(m_1)}^{\ell h}$ be the sorted virtual and
exact first-chunk scores. We define the empirical quantile map
\begin{equation}
\label{eq:calibration}
M_{\ell h}(\hat s_{(k)}^{\ell h})=s_{(k)}^{\ell h},
\qquad k=1,\ldots,m_1,
\end{equation}
with piecewise-linear interpolation between knots and clipping outside the
endpoint knots. The
exact first-chunk scores are retained. Each later virtual score is replaced by
$M_{\ell h}(\hat s_i^{\ell h})$.

We use a 2{,}048-token reconstruction-input budget. The exact first chunk
therefore contains at most $2{,}048-P_1$ context tokens so that it and its repeat
prompt fit within this budget. Later chunks can contain up to 2{,}048 context
tokens because virtual scoring does not process the prompt. If the context fits
in the exact first chunk, \method{} coincides with \kvzip{}. Otherwise, it
replaces \kvzip{}'s remaining
reconstruction passes with virtual scoring. Calibration is performed from scratch for
every request and uses no external examples, task labels, optimization, or
stored model-specific parameters.

\begin{figure}[htbp]
\centering
\includegraphics[width=0.85\linewidth]{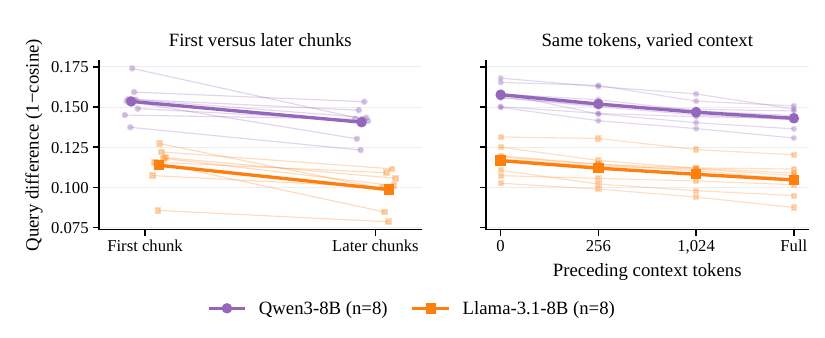}
\caption{Prefill-to-reconstruction query differences using
eight LongBench contexts per model. Thin lines show individual
contexts; thick lines show their mean. Left: the first chunk versus
the average of later full chunks. Right: fixed second-chunk tokens and
positions with varying preceding context, isolating the effect of full-context
conditioning. The vertical axis is the mean of one minus the cosine similarity
between matched prefill and reconstruction queries, so lower values indicate
closer states.}
\label{fig:first_chunk_mechanism}
\end{figure}

We choose the first chunk for exact reconstruction because its prefill states
have seen the least preceding context. Our diagnostic shows that the gap between prefill and reconstruction states is largest for the first chunk and shrinks in later chunks
(Figure~\ref{fig:first_chunk_mechanism}). The exact pass therefore provides
both high-quality scores for the first chunk and request-specific calibration data.
Appendix~\ref{app:first_chunk} gives measurement details and compares
within-chunk score rankings for the first and later chunks.

\subsection{Complexity analysis}
\label{sec:method:complexity}

We compare the compression work beyond standard prefill, ignoring
constant-size repeat prompts and sink sets. For $n\ge C$, the context contains
$\Theta(n/C)$ chunks. \kvzip{} reconstructs every chunk against the full
$n$-token cache, costing $O(LHdnC)$ per chunk. \method{} reconstructs only the
first chunk and scores all chunks virtually at $O(LHdC^2)$ per chunk.
The additional attention costs are therefore
\begin{align}
\operatorname{Cost}^{\mathrm{attn}}_{\mathrm{\kvzip}}
&=O(LHdn^2), \label{eq:kvzip_complexity}\\
\operatorname{Cost}^{\mathrm{attn}}_{\mathrm{\method}}
&=O(LHdnC). \label{eq:echo_complexity}
\end{align}
For fixed $C$, \method{} reduces compression attention work from quadratic to
linear in $n$, a factor of $n/C$. Standard prefill remains quadratic.
For model width $D$, projections and feed-forward layers add $O(LnD^2)$ for
\kvzip{} and $O(LCD^2)$ for \method{}.

Building and applying the calibration maps cost $O(LH_{kv}C\log C)$ and
$O(LH_{kv}n\log C)$, respectively. In both the KVPress and
\kvzip{} repositories, \method{} scores each layer during prefill and then
releases its queries, retaining only scalar importance scores across layers.

\section{Experiments}
\label{sec:experiments}

\subsection{Setup}
\label{sec:setup}

We evaluate Qwen3-8B \citep{yang2025qwen3} and Llama-3.1-8B-Instruct
\citep{grattafiori2024llama3}, abbreviated Llama-3.1-8B in tables and
figures. Accuracy experiments use KVPress
\citep{kvpress2024,devoto2025expectedattention} with bf16, SDPA attention, and batch size one. We compare
the uncompressed cache, \kvzip{}, and the \method{} configuration from
Section~\ref{sec:method}: a 2{,}048-token reconstruction-input budget, exact
first-chunk reconstruction, a separate virtual-to-exact quantile map for each
request, layer, and KV head, and global eviction of the lowest-scoring KV
pairs. The eviction ratio $r$ is the fraction of KV pairs removed; we test
$r\in\{0.5,0.75,0.9\}$ on LongBench and RULER-4K.

We evaluate 12 English LongBench tasks \citep{bai2024longbench} using
KVPress, truncating contexts to 32K tokens. RULER-4K \citep{hsieh2024ruler} contains 13 tasks
with 500 examples each and uses reference-string matching. Both benchmarks use
greedy decoding and per-task output limits. We report equally weighted task
averages from one run with seed 42.
The evaluator compresses each distinct context once without its questions,
then answers its questions independently from that shared cache.

We additionally evaluate RULER at 32K and 64K with 20 examples per task
across all 13 tasks, using identical inputs across methods within each model
and length. These runs compare the uncompressed cache and both compression
methods at $r\in\{0.75,0.9\}$. The task budgets are 32{,}768 and 65{,}536
tokens, including the generator's output allowance.
Appendix~\ref{app:long_context} details data generation and model settings.

For the latency evaluation, we use \kvzip{}'s
\href{https://github.com/snu-mllab/KVzip}{official CUDA implementation} and an
\method{} adapter built on the same codebase, using bf16, FlashAttention-2
\citep{dao2024flashattention2}, batch size one,
2{,}000-token context chunks, and one 80\,GB A100 GPU. Appendix~\ref{app:latency}
describes the prefill configurations.
Table~\ref{tab:protocol} (Appendix~\ref{app:latency}) summarizes the experimental
settings used to evaluate accuracy and timing for \kvzip{} and \method{}.

\subsection{Accuracy}
\label{sec:accuracy}

\begin{figure}[htbp]
\centering
\includegraphics[width=\linewidth]{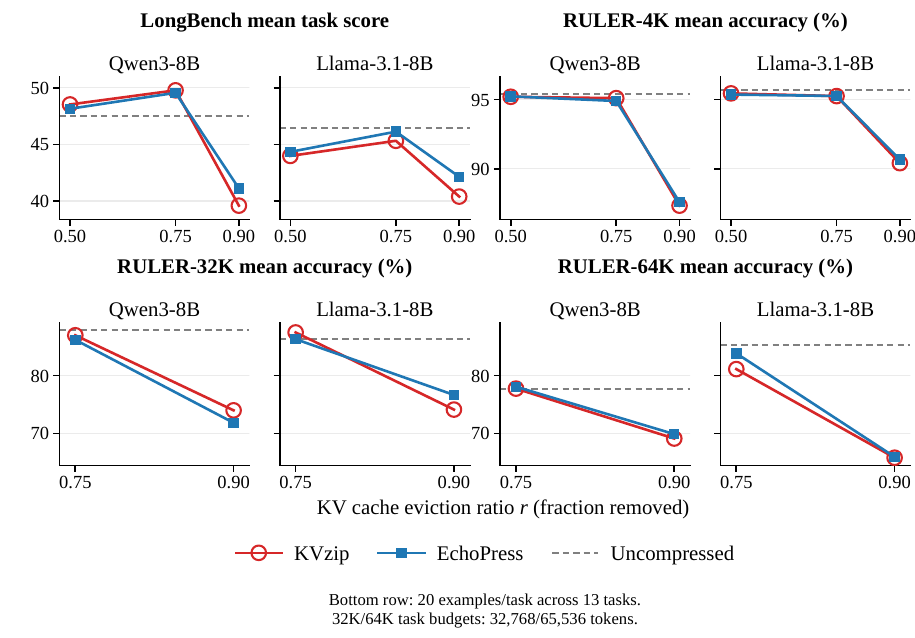}
\caption{Benchmark averages versus KV eviction ratio $r$ (higher is better).
The top row shows LongBench and RULER-4K; the bottom row shows RULER-32K/64K
with 20 examples per task. Dashed lines show uncompressed scores.
Additional results are in Appendix~\ref{app:accuracy}.}
\label{fig:benchmark_accuracy}
\end{figure}

Figure~\ref{fig:benchmark_accuracy} compares benchmark averages;
Tables~\ref{tab:longbench} and~\ref{tab:ruler} give category and task-group
means. On LongBench, \method{} is
0.38 and 0.22 points below \kvzip{} on Qwen at $r=0.5$ and $0.75$, while it is
0.36 and 0.81 points higher on Llama. At $r=0.9$, \method{} exceeds \kvzip{}
by 1.53 points on Qwen and 1.75 points on Llama.

On RULER-4K, \method{} and \kvzip{} differ by at most 0.21 points at
$r\leq0.75$. At $r=0.9$, \method{} is 0.24 points higher on Qwen and 0.26
points higher on Llama.

At 32K and 64K, \method{}'s mean accuracy ranges from 2.15 points below
to 2.72 points above \kvzip{} across both models and eviction ratios
(Figure~\ref{fig:benchmark_accuracy}, bottom row). At $r=0.9$, \method{} is
2.15 points below \kvzip{} on Qwen at 32K and 0.76 points above it at 64K.
On Llama, it is 2.53 and 0.15 points higher, respectively.
Table~\ref{tab:ruler_long_context} gives the mean scores from these
20-example-per-task evaluations.

To compare against a broader set of compression methods, we also run
\method{} at the KVPress leaderboard ratios $r\in\{0.25,0.5,0.75,0.875\}$.
Figure~\ref{fig:teaser}(a) overlays the Qwen3-8B runs on the public RULER-4K
leaderboard snapshot. At $r=0.875$, \method{} scores \RuQwenEchoSevenEighths{} on Qwen3-8B
and \RuLlamaEchoSevenEighths{} on Llama-3.1-8B-Instruct, compared with
\LbdKvzipQwenSevenEighths{} and \LbdKvzipLlamaSevenEighths{} for the
leaderboard's published \kvzip{} runs.

\subsection{Compression latency}
\label{sec:latency}

Figure~\ref{fig:teaser}(b) compares compression overhead for \kvzip{}'s official
CUDA implementation and our \method{} adapter at 4K--128K context tokens. We report medians over five repetitions
after one warm-up. Compression overhead is the median end-to-end time from the
start of prefill through physical compaction minus the median time of the
matching plain-prefill configuration. For
\method{}, this measurement includes the exact first-chunk forward, virtual
scoring, per-request sorting and interpolation, and compaction. The gap grows
with context length because \kvzip{} reconstructs every chunk, whereas
\method{} reconstructs only the first. Appendix~\ref{app:latency} reports the
implementation versions and the full compression-overhead sweep
(Table~\ref{tab:latency}).

\begin{figure}[htbp]
\centering
\includegraphics[width=0.82\linewidth]{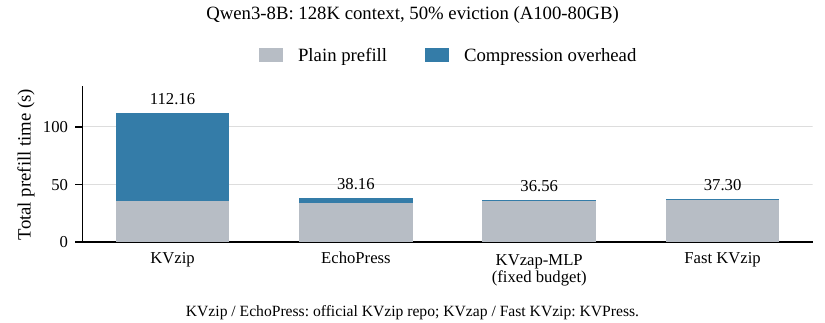}
\caption{Total prefill time for Qwen3-8B at 128K tokens and a target of 50\% eviction on one
80\,GB A100 GPU. \kvzip{} and \method{} use KVzip's official codebase with
FlashAttention-2; KVzap-MLP and Fast KVzip use KVPress/SDPA. Each bar stacks its
plain-prefill median and compression overhead, including physical compaction.
KVzap-MLP uses a fixed retained budget.}
\label{fig:latency_breakdown}
\end{figure}

At 128K tokens, \method{} reduces compression overhead from 77.02 to
3.94 seconds on Qwen3-8B, a 19.6$\times$ speedup, and from 67.31 to
3.52 seconds on Llama-3.1-8B-Instruct, a 19.1$\times$ speedup. Across the
complete 4K--128K sweep, speedup grows from 1.7$\times$ to 19.6$\times$.
Total prefill time (Table~\ref{tab:total_prefill}) falls
from 112.16 to 38.16 seconds on Qwen and from 98.26 to 34.02 seconds on Llama,
or 2.94$\times$ and 2.89$\times$.
The stage breakdown in Figure~\ref{fig:native_latency_breakdown}
(Appendix~\ref{app:latency}) attributes this gain to scoring and reconstruction.

Figure~\ref{fig:latency_breakdown} adds the training-based methods KVzap-MLP and Fast KVzip
\citep{jegou2026kvzap,kim2026fastkvzip} to this comparison at 128K tokens
and a target of 50\% eviction. We use their released learned scorers in KVPress,
adapting KVzap's threshold policy to a fixed cache budget. Their compression
overheads are 0.34 and 1.08 seconds, respectively, against a matching
plain-prefill median of 36.22 seconds. \method{} is training-free and achieves
a similar total prefill time in this comparison: 38.16 seconds, compared with
36.56 seconds for KVzap-MLP and 37.30 seconds for Fast KVzip.

The calibration operation itself is inexpensive. In a separate matched
KVPress timing ablation, constructing and applying a request-specific map takes
at most 1.24\,ms longer than applying a precomputed map. This changes 128K
compression overhead by less than 1\% on both models.

\subsection{Analysis and ablations}
\label{sec:analysis}

\begin{figure}[htbp]
\centering
\includegraphics[width=0.75\linewidth]{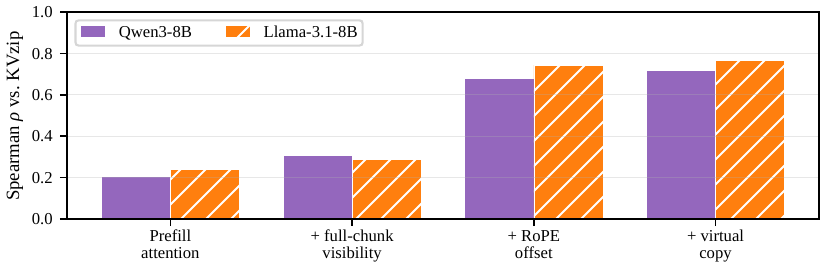}
\caption{Ablation of the virtual scorer, adding one component at a time. Each bar reports within-head rank agreement (Spearman correlation) with
exact reconstruction scores, computed as the mean over eight contexts of the
within-context median across layer--KV-head correlations.}
\label{fig:scorer_ablation}
\end{figure}

In Figure~\ref{fig:scorer_ablation}, we ablate the virtual scorer by adding one component at a time, evaluating on eight LongBench contexts per model. For each context and scorer, we
compute one Spearman correlation per layer and KV head across context positions,
take the median of those correlations, and then average the eight medians.
Exposing all keys in the current cached chunk gives a small improvement over
causal prefill attention, raising this statistic from 0.208 to 0.305 on Qwen and
from 0.237 to 0.288 on Llama.
Applying the RoPE position offset gives the largest gain, reaching 0.678 and
0.740, respectively. Adding the causal virtual-copy branch raises agreement
further to 0.720 and 0.768.

Table~\ref{tab:calibration_ablation} isolates calibration methods on a fixed 25\%
subset of RULER-4K at $r=0.9$. A single global per-request map is insufficient,
confirming that scale distortion varies by layer and KV head. Global eviction
compares the
exact scores retained for the first chunk with virtual scores from all later
chunks. We therefore leave the reliable exact scores unchanged and map each
later virtual score onto the exact-score scale. Reversing the map instead
distorts the exact first-chunk scores by moving them onto the less reliable
virtual scale. The proposed per-layer, per-KV-head virtual-to-exact map recovers
\kvzip{} accuracy and gives the best needle retrieval score in the table.

\begin{table}[htbp]
\centering
\caption{First-chunk calibration ablation on Qwen3-8B using a fixed 25\%
RULER-4K subset at $r=0.9$. Avg. weights all 13 tasks equally. NIAH weights
the eight needle-retrieval tasks equally.}
\label{tab:calibration_ablation}
\small
\setlength{\tabcolsep}{5pt}
\begin{tabular}{lrr}
\toprule
Scoring and calibration & RULER Avg. & NIAH (8) \\
\midrule
Exact first chunk, no calibration & 71.0 & 72.4 \\
Per-request, per-layer/head exact$\rightarrow$virtual & 50.8 & 43.8 \\
Per-request, global virtual$\rightarrow$exact & 77.3 & 83.5 \\
\method{}: per-request, per-layer/head virtual$\rightarrow$exact & \textbf{87.6} & \textbf{95.7} \\
\kvzip{} & 87.5 & 92.5 \\
\bottomrule
\end{tabular}

\end{table}

\section{Related work}
\label{sec:related}

KV eviction uses attention scores
\citep{zhang2023h2o,oren2024tova,li2024snapkv,liu2023scissorhands,chen2024nacl},
layer/head budgets \citep{cai2024pyramidkv,feng2025adakv,yang2024pyramidinfer},
or initial and recent tokens \citep{xiao2024streamingllm}.
FastGen profiles prompts to select head policies; DuoAttention optimizes
which heads need full context \citep{ge2024fastgen,xiao2025duoattention}.
FINCH conditions eviction on a prompt; task-aware compression uses task
descriptions or examples \citep{corallo2024finch,corallo2025takc}.
Query-agnostic scores use key geometry or expected attention
\citep{devoto2024knorm,park2025keydiff,godey2025qfilters,devoto2025expectedattention}.
Compactor combines non-causal attention and leverage scores
\citep{chari2025compactor}. InfiniPot uses auxiliary prompts and novelty scores;
EpiCache compresses conversational episodes \citep{kim2024infinipot,kim2026epicache}.
LU-KV uses offline head utilities \citep{tang2026lukv}.

Learned retention includes Locret, DMS, and TRIM-KV
\citep{huang2024locret,lancucki2025dms,bui2025trimkv}.
KVzap and Fast KVzip replace \kvzip{}'s reconstruction scoring with
learned predictors or gates
\citep{jegou2026kvzap,kim2026fastkvzip}.
RestoreKV learns cache augmentation before eviction \citep{baek2026restorekv}.
\method{} approximates reconstruction from prefill states and calibrates per
request without training. CacheBlend and EPIC inspired our reuse of states
under changed context \citep{yao2025cacheblend,hu2025epic}.

Other approaches prune key channels \citep{xu2025think}, quantize KV entries
\citep{liu2024kivi,yang2024mikv,he2024zipcache}, sparsify attention computation
\citep{tang2024quest,jiang2024minference,gao2024seerattention,lu2025moba}, or
selectively access offloaded caches
\citep{lee2024infinigen,liu2024retrievalattention,chen2025magicpig}.
These change entry size or cache access; \method{} permanently removes
entries. Deja Vu studies contextual sparsity; LLM-Microscope examines
information held by individual tokens
\citep{liu2023dejavu,razzhigaev2025microscope}. Our analysis examines which
reconstruction computations can be approximated from prefill states.

\section{Discussion and limitations}
\label{sec:discussion}

\method{} compresses the prefilled context but performs no further eviction
during decoding, so the cache grows with generated tokens. Prior work manages
this growth through eviction \citep{zhang2023h2o,oren2024tova,anagnostidis2023dynamic,cai2025rkv,du2025rlkv}
or learned compressed memory \citep{kim2024ccm}.
Extending our scorer to completed generation chunks would require buffering
query states and choosing a scoring frequency that balances buffering cost,
cache savings, and latency. We leave this extension and its evaluation to
future work.

Benchmark averages also hide task-level regressions: at $r\geq0.75$,
\method{} trails \kvzip{} on LongBench code completion on both models, and at
$r=0.9$ on RULER common-word extraction for Qwen
(Tables~\ref{tab:longbench_full}--\ref{tab:ruler_full_llama}).
\section{Conclusion}
\label{sec:conclusion}

We introduced \method{}, a training-free method for query-agnostic KV cache
compression. It reuses prefill states to approximate reconstruction scores and
calibrates later scores per layer and KV head using one exact first-chunk
reconstruction. On LongBench and RULER-4K with Qwen3-8B and
Llama-3.1-8B-Instruct, it closely matches \kvzip{} at moderate eviction and
exceeds it at 90\%. RULER-32K/64K means stay within 2.8 points of \kvzip{}. At 128K tokens,
it reduces compression overhead by 19.1--19.6$\times$ on an 80\,GB A100 GPU
across both models.

\bibliography{references}
\bibliographystyle{plainnat}

\appendix
\clearpage
\raggedbottom
\setlength{\intextsep}{8pt plus 2pt minus 2pt}

\begingroup
\setlength{\intextsep}{6pt plus 2pt minus 2pt}
\setlength{\abovecaptionskip}{6pt}
\renewcommand{\arraystretch}{0.95}

\section{Additional accuracy results}
\label{app:accuracy}

Tables~\ref{tab:longbench}--\ref{tab:ruler_full_llama} expand the results of
Section~\ref{sec:accuracy} using the settings in Section~\ref{sec:setup}.
Here $r$ is the fraction of KV pairs evicted.

\begin{table}[H]
\centering
\small
\caption{LongBench category means. Avg. weights all 12 tasks equally. Retr.
denotes passage retrieval and Code denotes LCC. Bold marks the better compressed method at each $r$ and displayed
precision unless tied.}
\label{tab:longbench}
\setlength{\tabcolsep}{3pt}
\begin{tabular}{@{}lcccccccc@{}}
\toprule
\multicolumn{9}{l}{\textit{Qwen3-8B}} \\
Method & $r$ & \shortstack{Single\\QA} & \shortstack{Multi\\QA} & Summ. & \shortstack{Few-\\shot} & Retr. & Code & Avg. \\
\midrule
Uncompressed & 0 & 42.4 & 49.2 & 29.1 & 40.7 & 91.4 & 64.5 & 47.5 \\
\addlinespace[2pt]
KVzip & 0.5 & 42.4 & 48.4 & \textbf{29.2} & \textbf{46.4} & 94.0 & \textbf{64.7} & \textbf{48.5} \\
EchoPress & 0.5 & 42.4 & \textbf{48.7} & 29.1 & 44.0 & \textbf{95.0} & 63.4 & 48.2 \\
\addlinespace[2pt]
KVzip & 0.75 & 41.7 & 49.0 & 28.7 & 56.5 & 92.0 & \textbf{63.1} & \textbf{49.8} \\
EchoPress & 0.75 & \textbf{42.1} & \textbf{50.0} & \textbf{29.0} & \textbf{56.6} & \textbf{93.6} & 54.0 & 49.6 \\
\addlinespace[2pt]
KVzip & 0.9 & \textbf{37.1} & 38.5 & 26.8 & 46.9 & 54.2 & \textbf{47.0} & 39.6 \\
EchoPress & 0.9 & 35.9 & \textbf{39.3} & \textbf{27.1} & \textbf{49.8} & \textbf{76.5} & 37.6 & \textbf{41.1} \\
\midrule
\multicolumn{9}{l}{\textit{Llama-3.1-8B}} \\
Method & $r$ & \shortstack{Single\\QA} & \shortstack{Multi\\QA} & Summ. & \shortstack{Few-\\shot} & Retr. & Code & Avg. \\
\midrule
Uncompressed & 0 & 43.8 & 48.0 & 30.1 & 35.2 & 100.0 & 52.0 & 46.5 \\
\addlinespace[2pt]
KVzip & 0.5 & 44.4 & 44.2 & \textbf{30.1} & 25.4 & 100.0 & \textbf{50.9} & 44.0 \\
EchoPress & 0.5 & 44.4 & \textbf{45.7} & 30.0 & \textbf{25.9} & 100.0 & 50.2 & \textbf{44.3} \\
\addlinespace[2pt]
KVzip & 0.75 & 42.4 & \textbf{43.3} & \textbf{29.8} & 39.8 & 98.5 & \textbf{49.0} & 45.3 \\
EchoPress & 0.75 & \textbf{43.8} & 42.8 & 29.5 & \textbf{45.8} & \textbf{100.0} & 43.2 & \textbf{46.1} \\
\addlinespace[2pt]
KVzip & 0.9 & \textbf{42.0} & 38.7 & 27.8 & 42.0 & 61.5 & \textbf{41.3} & 40.4 \\
EchoPress & 0.9 & 40.2 & \textbf{42.4} & 27.8 & \textbf{45.2} & \textbf{81.5} & 30.3 & \textbf{42.1} \\
\bottomrule
\end{tabular}

\end{table}

\begin{table}[H]
\centering
\small
\caption{RULER-4K group means. NIAH denotes eight needle-retrieval tasks, VT
variable tracking, CWE/FWE common/frequent-word extraction, and QA the two
question-answering tasks. Avg. weights all 13 tasks equally.}
\label{tab:ruler}
\setlength{\tabcolsep}{4pt}
\begin{tabular}{@{}lccccccc@{}}
\toprule
\multicolumn{8}{l}{\textit{Qwen3-8B}} \\
Method & $r$ & NIAH (8) & VT & CWE & FWE & QA (2) & Avg. \\
\midrule
Uncompressed & 0 & 100.0 & 100.0 & 98.9 & 95.4 & 72.9 & 95.4 \\
\addlinespace[2pt]
KVzip & 0.5 & 100.0 & 100.0 & 99.0 & \textbf{95.9} & 71.4 & 95.2 \\
EchoPress & 0.5 & 100.0 & 100.0 & \textbf{99.1} & 95.7 & \textbf{71.7} & 95.2 \\
\addlinespace[2pt]
KVzip & 0.75 & \textbf{99.9} & 100.0 & \textbf{98.9} & \textbf{96.4} & \textbf{70.8} & \textbf{95.1} \\
EchoPress & 0.75 & 99.8 & 100.0 & 98.0 & 95.9 & 70.5 & 94.9 \\
\addlinespace[2pt]
KVzip & 0.9 & 92.8 & 100.0 & \textbf{78.2} & \textbf{84.6} & 64.9 & 87.3 \\
EchoPress & 0.9 & \textbf{95.5} & 100.0 & 59.4 & 84.3 & \textbf{65.5} & \textbf{87.6} \\
\midrule
\multicolumn{8}{l}{\textit{Llama-3.1-8B}} \\
Method & $r$ & NIAH (8) & VT & CWE & FWE & QA (2) & Avg. \\
\midrule
Uncompressed & 0 & 99.9 & 99.9 & 99.6 & 94.8 & 75.1 & 95.7 \\
\addlinespace[2pt]
KVzip & 0.5 & 99.9 & 99.9 & 99.7 & 94.1 & \textbf{73.8} & \textbf{95.5} \\
EchoPress & 0.5 & 99.9 & 99.9 & 99.7 & \textbf{94.4} & 73.1 & 95.4 \\
\addlinespace[2pt]
KVzip & 0.75 & 100.0 & \textbf{99.9} & \textbf{98.7} & \textbf{94.5} & 72.8 & 95.3 \\
EchoPress & 0.75 & 100.0 & 99.8 & 98.2 & 93.9 & \textbf{73.4} & 95.3 \\
\addlinespace[2pt]
KVzip & 0.9 & \textbf{98.1} & \textbf{99.3} & 69.4 & 89.9 & 66.0 & 90.4 \\
EchoPress & 0.9 & 98.0 & 98.9 & \textbf{70.9} & 89.9 & \textbf{67.4} & \textbf{90.7} \\
\bottomrule
\end{tabular}

\end{table}

\subsection{Per-task results}
\label{app:pertask}

LongBench uses task-specific metrics and contexts truncated to 32K tokens;
RULER-4K uses string-match accuracy on 500 examples per task.

\begin{table}[H]
\centering
\small
\caption{LongBench per-task scores on Qwen3-8B.}
\label{tab:longbench_full}
\setlength{\tabcolsep}{3pt}
\begin{tabular}{@{}lccccccc@{}}
\toprule
\multicolumn{8}{l}{\textit{Question answering}} \\
Method & $r$ & Qasper & MultiFQA & NrtvQA & HotpotQA & 2WikiMQA & MuSiQue \\
\midrule
Uncompressed & 0 & 44.8 & 55.6 & 26.7 & 62.9 & 49.2 & 35.4 \\
\addlinespace[2pt]
KVzip & 0.5 & 44.5 & 55.8 & \textbf{26.8} & 62.8 & 47.8 & \textbf{34.7} \\
EchoPress & 0.5 & \textbf{44.6} & \textbf{56.4} & 26.1 & \textbf{62.9} & \textbf{48.8} & 34.5 \\
\addlinespace[2pt]
KVzip & 0.75 & 42.6 & \textbf{55.0} & 27.6 & \textbf{63.8} & 47.8 & 35.3 \\
EchoPress & 0.75 & \textbf{43.4} & 54.4 & \textbf{28.4} & 63.7 & \textbf{48.9} & \textbf{37.3} \\
\addlinespace[2pt]
KVzip & 0.9 & \textbf{36.9} & \textbf{51.0} & 23.2 & 53.3 & 37.7 & 24.4 \\
EchoPress & 0.9 & 35.1 & 48.0 & \textbf{24.5} & \textbf{53.8} & \textbf{39.6} & \textbf{24.5} \\
\bottomrule
\end{tabular}

\par\medskip
\begin{tabular}{@{}lccccccc@{}}
\toprule
\multicolumn{8}{l}{\textit{Other tasks}} \\
Method & $r$ & GovReport & QMSum & TREC & SAMSum & PsgRetr & LCC \\
\midrule
Uncompressed & 0 & 33.6 & 24.5 & 41.0 & 40.3 & 91.4 & 64.5 \\
\addlinespace[2pt]
KVzip & 0.5 & \textbf{33.8} & \textbf{24.7} & \textbf{52.5} & 40.4 & 94.0 & \textbf{64.7} \\
EchoPress & 0.5 & 33.6 & 24.6 & 47.0 & \textbf{41.0} & \textbf{95.0} & 63.4 \\
\addlinespace[2pt]
KVzip & 0.75 & 33.4 & 24.0 & 72.5 & \textbf{40.6} & 92.0 & \textbf{63.1} \\
EchoPress & 0.75 & \textbf{33.7} & \textbf{24.3} & \textbf{73.0} & 40.1 & \textbf{93.6} & 54.0 \\
\addlinespace[2pt]
KVzip & 0.9 & 30.9 & 22.6 & 57.0 & 36.7 & 54.2 & \textbf{47.0} \\
EchoPress & 0.9 & \textbf{31.4} & \textbf{22.8} & \textbf{60.8} & \textbf{38.8} & \textbf{76.5} & 37.6 \\
\bottomrule
\end{tabular}

\end{table}

\begin{table}[H]
\centering
\small
\caption{LongBench per-task scores on Llama-3.1-8B-Instruct.}
\label{tab:longbench_full_llama}
\setlength{\tabcolsep}{3pt}
\begin{tabular}{@{}lccccccc@{}}
\toprule
\multicolumn{8}{l}{\textit{Question answering}} \\
Method & $r$ & Qasper & MultiFQA & NrtvQA & HotpotQA & 2WikiMQA & MuSiQue \\
\midrule
Uncompressed & 0 & 47.1 & 55.3 & 28.9 & 59.5 & 51.8 & 32.7 \\
\addlinespace[2pt]
KVzip & 0.5 & 47.3 & 56.8 & \textbf{29.1} & 55.8 & 48.2 & 28.6 \\
EchoPress & 0.5 & 47.3 & \textbf{57.0} & 28.7 & \textbf{57.4} & \textbf{49.7} & \textbf{29.9} \\
\addlinespace[2pt]
KVzip & 0.75 & 44.6 & 54.8 & 27.9 & \textbf{56.1} & \textbf{46.4} & 27.4 \\
EchoPress & 0.75 & \textbf{46.0} & \textbf{55.6} & \textbf{29.7} & 54.2 & 46.3 & \textbf{28.0} \\
\addlinespace[2pt]
KVzip & 0.9 & \textbf{39.8} & \textbf{55.5} & \textbf{30.8} & 51.7 & 39.4 & 25.1 \\
EchoPress & 0.9 & 39.4 & 53.1 & 28.0 & \textbf{55.5} & \textbf{46.1} & \textbf{25.6} \\
\bottomrule
\end{tabular}

\par\medskip
\begin{tabular}{@{}lccccccc@{}}
\toprule
\multicolumn{8}{l}{\textit{Other tasks}} \\
Method & $r$ & GovReport & QMSum & TREC & SAMSum & PsgRetr & LCC \\
\midrule
Uncompressed & 0 & 35.3 & 24.9 & 29.5 & 40.9 & 100.0 & 52.0 \\
\addlinespace[2pt]
KVzip & 0.5 & \textbf{34.8} & \textbf{25.4} & \textbf{16.5} & 34.3 & 100.0 & \textbf{50.9} \\
EchoPress & 0.5 & 34.7 & 25.3 & 15.5 & \textbf{36.4} & 100.0 & 50.2 \\
\addlinespace[2pt]
KVzip & 0.75 & \textbf{34.4} & \textbf{25.1} & 47.0 & 32.6 & 98.5 & \textbf{49.0} \\
EchoPress & 0.75 & 34.1 & 24.9 & \textbf{59.0} & 32.6 & \textbf{100.0} & 43.2 \\
\addlinespace[2pt]
KVzip & 0.9 & \textbf{31.8} & 23.9 & 45.5 & \textbf{38.5} & 61.5 & \textbf{41.3} \\
EchoPress & 0.9 & 31.5 & \textbf{24.2} & \textbf{52.5} & 38.0 & \textbf{81.5} & 30.3 \\
\bottomrule
\end{tabular}

\end{table}

\begin{table}[H]
\centering
\small
\caption{RULER-4K per-task accuracy on Qwen3-8B. S1--S3 and MK1--MK3 are
single-needle and multi-key variants. MQ/MV are multi-query/value retrieval.}
\label{tab:ruler_full}
\setlength{\tabcolsep}{3pt}
\begin{tabular}{@{}lccccccccc@{}}
\toprule
\multicolumn{10}{l}{\textit{Needle retrieval}} \\
Method & $r$ & S1 & S2 & S3 & MK1 & MK2 & MK3 & MQ & MV \\
\midrule
Uncompressed & 0 & 100.0 & 100.0 & 100.0 & 100.0 & 100.0 & 100.0 & 99.9 & 100.0 \\
\addlinespace[2pt]
KVzip & 0.5 & 100.0 & 100.0 & 100.0 & 100.0 & 100.0 & 100.0 & 100.0 & 100.0 \\
EchoPress & 0.5 & 100.0 & 100.0 & 100.0 & 100.0 & 100.0 & 100.0 & 100.0 & 100.0 \\
\addlinespace[2pt]
KVzip & 0.75 & 100.0 & 100.0 & 100.0 & \textbf{99.8} & 100.0 & \textbf{99.8} & 100.0 & 99.9 \\
EchoPress & 0.75 & 100.0 & 100.0 & 100.0 & 99.6 & 100.0 & 99.2 & 100.0 & \textbf{100.0} \\
\addlinespace[2pt]
KVzip & 0.9 & 100.0 & 95.4 & 98.4 & 83.0 & 97.2 & \textbf{92.0} & 97.8 & 79.0 \\
EchoPress & 0.9 & 100.0 & \textbf{99.0} & \textbf{99.0} & \textbf{91.6} & \textbf{98.0} & 90.6 & \textbf{98.0} & \textbf{87.8} \\
\bottomrule
\end{tabular}

\par\medskip
\begin{tabular}{@{}lcccccc@{}}
\toprule
\multicolumn{7}{l}{\textit{Other tasks}} \\
Method & $r$ & VT & CWE & FWE & QA1 & QA2 \\
\midrule
Uncompressed & 0 & 100.0 & 98.9 & 95.4 & 82.0 & 63.8 \\
\addlinespace[2pt]
KVzip & 0.5 & 100.0 & 99.0 & \textbf{95.9} & 80.6 & 62.2 \\
EchoPress & 0.5 & 100.0 & \textbf{99.1} & 95.7 & 80.6 & \textbf{62.8} \\
\addlinespace[2pt]
KVzip & 0.75 & 100.0 & \textbf{98.9} & \textbf{96.4} & \textbf{81.0} & \textbf{60.6} \\
EchoPress & 0.75 & 100.0 & 98.0 & 95.9 & 80.6 & 60.4 \\
\addlinespace[2pt]
KVzip & 0.9 & 100.0 & \textbf{78.2} & \textbf{84.6} & \textbf{74.4} & 55.4 \\
EchoPress & 0.9 & 100.0 & 59.4 & 84.3 & 74.2 & \textbf{56.8} \\
\bottomrule
\end{tabular}

\end{table}

\begin{table}[H]
\centering
\small
\caption{RULER-4K per-task accuracy on Llama-3.1-8B-Instruct.}
\label{tab:ruler_full_llama}
\setlength{\tabcolsep}{3pt}
\begin{tabular}{@{}lccccccccc@{}}
\toprule
\multicolumn{10}{l}{\textit{Needle retrieval}} \\
Method & $r$ & S1 & S2 & S3 & MK1 & MK2 & MK3 & MQ & MV \\
\midrule
Uncompressed & 0 & 100.0 & 100.0 & 99.8 & 99.8 & 100.0 & 99.8 & 99.9 & 99.9 \\
\addlinespace[2pt]
KVzip & 0.5 & 99.8 & 100.0 & 100.0 & 100.0 & 100.0 & 99.8 & 100.0 & 100.0 \\
EchoPress & 0.5 & 99.8 & 100.0 & 100.0 & 100.0 & 100.0 & 99.8 & 100.0 & 100.0 \\
\addlinespace[2pt]
KVzip & 0.75 & 100.0 & 100.0 & 99.8 & 100.0 & 100.0 & \textbf{100.0} & 99.9 & 99.9 \\
EchoPress & 0.75 & 100.0 & 100.0 & \textbf{100.0} & 100.0 & 100.0 & 99.8 & \textbf{100.0} & \textbf{100.0} \\
\addlinespace[2pt]
KVzip & 0.9 & 100.0 & 99.8 & \textbf{100.0} & 98.8 & \textbf{98.6} & 95.6 & \textbf{99.1} & 92.8 \\
EchoPress & 0.9 & 100.0 & \textbf{100.0} & 96.4 & \textbf{99.8} & 98.2 & \textbf{97.0} & 98.3 & \textbf{94.5} \\
\bottomrule
\end{tabular}

\par\medskip
\begin{tabular}{@{}lcccccc@{}}
\toprule
\multicolumn{7}{l}{\textit{Other tasks}} \\
Method & $r$ & VT & CWE & FWE & QA1 & QA2 \\
\midrule
Uncompressed & 0 & 99.9 & 99.6 & 94.8 & 87.8 & 62.4 \\
\addlinespace[2pt]
KVzip & 0.5 & 99.9 & 99.7 & 94.1 & \textbf{85.6} & \textbf{62.0} \\
EchoPress & 0.5 & 99.9 & 99.7 & \textbf{94.4} & 85.4 & 60.8 \\
\addlinespace[2pt]
KVzip & 0.75 & \textbf{99.9} & \textbf{98.7} & \textbf{94.5} & 84.8 & 60.8 \\
EchoPress & 0.75 & 99.8 & 98.2 & 93.9 & \textbf{85.6} & \textbf{61.2} \\
\addlinespace[2pt]
KVzip & 0.9 & \textbf{99.3} & 69.4 & 89.9 & 77.6 & 54.4 \\
EchoPress & 0.9 & 98.9 & \textbf{70.9} & 89.9 & \textbf{78.6} & \textbf{56.2} \\
\bottomrule
\end{tabular}

\end{table}

\subsection{RULER at 32K and 64K}
\label{app:long_context}

Table~\ref{tab:ruler_long_context} reports equally weighted means over 13
RULER tasks. At each model and length, all five configurations share the same
260 examples (20 per task): uncompressed, \kvzip{}, and \method{}, with
$r\in\{0.75,0.9\}$ for compression. Table~\ref{tab:long_context_settings}
lists the evaluation settings.

\begin{table}[H]
\centering
\small
\caption{RULER accuracy at 32K and 64K, using 20 examples per task. Full denotes
the uncompressed baseline. $\Delta$ is \method{} minus \kvzip{} in percentage
points.}
\label{tab:ruler_long_context}
\setlength{\tabcolsep}{5pt}
\begin{tabular}{@{}lccccccccc@{}}
\toprule
 & & \multicolumn{4}{c}{Qwen3-8B} & \multicolumn{4}{c}{Llama-3.1-8B} \\
\cmidrule(lr){3-6}\cmidrule(l){7-10}
Budget & $r$ & Full & \kvzip{} & \method{} & $\Delta$ & Full & \kvzip{} & \method{} & $\Delta$ \\
\midrule
32K & 0.75 & 87.88 & 86.98 & 86.19 & $-0.79$ & 86.38 & 87.47 & 86.31 & $-1.17$ \\
32K & 0.9 & 87.88 & 73.99 & 71.83 & $-2.15$ & 86.38 & 74.12 & 76.65 & $+2.53$ \\
64K & 0.75 & 77.72 & 77.75 & 78.08 & $+0.33$ & 85.22 & 81.13 & 83.85 & $+2.72$ \\
64K & 0.9 & 77.72 & 69.11 & 69.87 & $+0.76$ & 85.22 & 65.79 & 65.94 & $+0.15$ \\
\bottomrule
\end{tabular}

\end{table}

\begin{table}[H]
\centering
\small
\caption{Settings for the RULER-32K/64K evaluation.}
\label{tab:long_context_settings}
\setlength{\tabcolsep}{4pt}
\begin{tabular}{@{}p{0.19\linewidth}p{0.77\linewidth}@{}}
\toprule
Setting & Configuration \\
\midrule
Data & NVIDIA/RULER v1 (\texttt{c3f5e3b4f87f}), base template, seed 42,
model-specific tokenizers; no truncation. \\
Lengths & Budgets of 32,768 / 65,536 tokens include output allowances;
actual prefill prefixes span 28,057--32,718 / 60,224--65,478 tokens. \\
Runtime & bf16, SDPA, batch size one; PyTorch 2.8.0 (CUDA 12.8),
transformers 5.2.0, Triton 3.4.0. \\
Positions & Qwen3: fixed YaRN factor four \citep{peng2024yarn}, original
context 32,768, checkpoint RoPE base. Llama: released RoPE configuration.
Both use a maximum position limit of 131,072. \\
Decoding & Greedy with official per-task output limits; Qwen thinking disabled. \\
Compression & The 2,048-token accuracy configuration, global KV-pair allocation,
and compression before the question. Question and generated tokens remain
uncompressed. \method{} uses exact first-chunk scores and per-request,
per-layer/KV-head calibration, without an offline map. \\
\bottomrule
\end{tabular}
\end{table}

RULER-4K uses the published KVPress dataset with a common Llama tokenizer,
so it is a separate evaluation, not a controlled point on this sweep.

\endgroup

\section{Compression latency}
\label{app:latency}

Tables~\ref{tab:latency} and~\ref{tab:total_prefill} compare three configurations
at 50\% eviction in the \href{https://github.com/snu-mllab/KVzip}{\kvzip{}
repository} (commit \texttt{\BenchNativeCommit{}}). The unmodified
``16K-chunk'' configuration prefills in 16{,}000-token chunks; ``full-prefill''
changes only this setting. \method{} scores virtually during full-context
prefill, releasing queries after each layer and retaining importance scores;
exact first-chunk reconstruction, calibration, and compaction follow.

\begin{table}[H]
\centering
\caption{Accuracy and timing settings for \kvzip{} and \method{}. Both use
bf16 and batch size one; timing uses one 80\,GB A100 at 50\% eviction.}
\label{tab:protocol}
\small
\setlength{\tabcolsep}{4pt}
\begin{tabular}{@{}p{0.17\linewidth}p{0.40\linewidth}p{0.37\linewidth}@{}}
\toprule
Setting & Accuracy & Timing \\
\midrule
Engine / attention & KVPress / SDPA & KVzip / FlashAttention-2 \\
Chunk budget & First chunk: 2,048 tokens including prompt; later chunks: up to 2,048 & 2,000 context tokens plus repeat prompt \\
Prompts / prefix & KVPress chat/prefix settings & KVzip prompts and prefix/sinks \\
Context coverage & RULER 4K/32K/64K; LongBench $\leq$32K & 4K--128K, excluding protected prefix \\
Output & Task scores & Compacted cache; no generation \\
\bottomrule
\end{tabular}
\end{table}

Timing uses FlashAttention-2~\BenchFlashAttnQwen{}, PyTorch~\BenchTorchQwen{}
(CUDA~\BenchCudaQwen{}), transformers~\BenchTransformersQwen{}, and
Triton~\BenchTritonQwen{}, with Qwen3 YaRN factor four. Each model uses the
same saved LongBench inputs across methods and five repetitions after warm-up. CUDA is synchronized at stage boundaries. Prompt preparation,
calibration, and compaction are included; loading, tokenization, compilation,
warm-up, verification, and cleanup are excluded. A 4K check confirms finite
scores/logits and bitwise-identical exact first-chunk scores, without testing
downstream accuracy equivalence to the accuracy configuration
(Section~\ref{sec:method:scale}).

Total prefill time is the median per-run duration through physical compaction,
excluding first-token generation. Compression overhead subtracts each
configuration's matching plain-prefill median. The separate \kvzip{} and
\method{} sweeps retain separate full-prefill references. Total medians use
raw repetitions, not sums of independently computed stage medians.

\begin{table}[H]
\centering
\caption{Plain-prefill baselines and compression overhead in seconds. Speedup
is \kvzip{} 16K-chunk overhead divided by \method{} overhead. ``Full K'' and
``Full E'' are the respective sweeps' full-prefill baselines.}
\label{tab:latency}
\small
\setlength{\tabcolsep}{4pt}
\begin{tabular}{@{}lccccccc@{}}
\toprule
\multicolumn{8}{l}{\textit{Qwen3-8B}} \\
Context & \shortstack{Plain\\16K-chunk} & \shortstack{Plain\\Full K} & \shortstack{Plain\\Full E} & \shortstack{KVzip\\16K-chunk} & \shortstack{KVzip\\full prefill} & EchoPress & Speedup \\
\midrule
4K & 0.36 & 0.36 & 0.36 & 0.77 & 0.78 & 0.46 & \textbf{1.7$\times$} \\
8K & 0.73 & 0.73 & 0.72 & 1.63 & 1.62 & 0.57 & \textbf{2.9$\times$} \\
16K & 1.67 & 1.63 & 1.61 & 3.68 & 3.68 & 0.79 & \textbf{4.7$\times$} \\
32K & 4.11 & 3.96 & 3.91 & 9.03 & 9.04 & 1.23 & \textbf{7.3$\times$} \\
64K & 11.23 & 10.90 & 10.78 & 24.53 & 24.47 & 2.14 & \textbf{11.5$\times$} \\
128K & 35.14 & 34.52 & 34.23 & 77.02 & 77.00 & 3.94 & \textbf{19.6$\times$} \\
\midrule
\multicolumn{8}{l}{\textit{Llama-3.1-8B}} \\
Context & \shortstack{Plain\\16K-chunk} & \shortstack{Plain\\Full K} & \shortstack{Plain\\Full E} & \shortstack{KVzip\\16K-chunk} & \shortstack{KVzip\\full prefill} & EchoPress & Speedup \\
\midrule
4K & 0.32 & 0.32 & 0.32 & 0.68 & 0.69 & 0.41 & \textbf{1.7$\times$} \\
8K & 0.66 & 0.66 & 0.66 & 1.43 & 1.43 & 0.51 & \textbf{2.8$\times$} \\
16K & 1.50 & 1.46 & 1.46 & 3.23 & 3.23 & 0.70 & \textbf{4.6$\times$} \\
32K & 3.67 & 3.55 & 3.54 & 7.93 & 7.92 & 1.09 & \textbf{7.2$\times$} \\
64K & 9.96 & 9.68 & 9.70 & 21.65 & 21.67 & 1.87 & \textbf{11.6$\times$} \\
128K & 30.95 & 30.36 & 30.50 & 67.31 & 67.31 & 3.52 & \textbf{19.1$\times$} \\
\bottomrule
\end{tabular}

\end{table}

\begin{table}[H]
\centering
\caption{Median total prefill time (seconds), including compression and
physical compaction, across five repetitions.}
\label{tab:total_prefill}
\small
\setlength{\tabcolsep}{5pt}
\begin{tabular}{@{}lrrrrrr@{}}
\toprule
 & \multicolumn{3}{c}{Qwen3-8B} & \multicolumn{3}{c}{Llama-3.1-8B} \\
\cmidrule(lr){2-4}\cmidrule(l){5-7}
Context & \shortstack{KVzip\\16K-chunk} & \shortstack{KVzip\\full} & EchoPress & \shortstack{KVzip\\16K-chunk} & \shortstack{KVzip\\full} & EchoPress \\
\midrule
4K & 1.13 & 1.14 & 0.81 & 1.01 & 1.01 & 0.74 \\
8K & 2.36 & 2.36 & 1.29 & 2.09 & 2.09 & 1.17 \\
16K & 5.34 & 5.31 & 2.39 & 4.73 & 4.69 & 2.17 \\
32K & 13.14 & 12.99 & 5.15 & 11.60 & 11.47 & 4.64 \\
64K & 35.76 & 35.37 & 12.92 & 31.61 & 31.35 & 11.57 \\
128K & 112.16 & 111.51 & 38.16 & 98.26 & 97.67 & 34.02 \\
\bottomrule
\end{tabular}

\end{table}

\begin{figure}[H]
\centering
\includegraphics[width=0.82\linewidth]{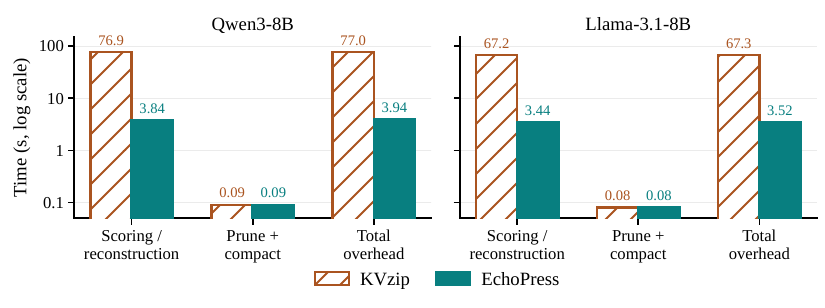}
\caption{Compression-time breakdown at 128K on one 80\,GB A100.
For \method{}, scoring includes exact first-chunk reconstruction, virtual
scoring, and calibration. Overhead subtracts the matching plain-prefill baseline.}
\label{fig:native_latency_breakdown}
\end{figure}

\subsection{Timing comparison with learned methods}
\label{app:trained_latency}

Figure~\ref{fig:latency_breakdown} combines the \kvzip{} 16K-chunk and \method{} timings above with
KVPress/SDPA timings of KVzap-MLP and Fast KVzip
(Table~\ref{tab:learned_timing_protocol}). All use Qwen3-8B, the same
131{,}072 context tokens, batch size one, and a 50\% eviction target on an
A100-SXM4-80GB. Separate sweeps use matching plain-prefill baselines. The
\kvzip{}/\method{} runs use a 28-token protected prefix and retain
0.49973/0.49992 of KV pairs; the learned methods retain exactly 0.5. Backends,
prompts, chunking, and compaction differ, so the figure compares measured
implementations. Training costs and retained-cache accuracy are not evaluated.

\begin{table}[H]
\centering
\caption{Settings for the KVPress implementations of the learned methods.}
\label{tab:learned_timing_protocol}
\small
\setlength{\tabcolsep}{4pt}
\begin{tabular}{@{}p{0.18\linewidth}p{0.78\linewidth}@{}}
\toprule
Setting & Configuration \\
\midrule
Input / prefill & Saved concatenated LongBench \texttt{gov\_report} contexts;
three-token chat prefix; 131,075 total tokens; one full-context prefill. \\
Software & PyTorch 2.8.0 (CUDA 12.8), transformers 5.2.0, Triton 3.4.0; SDPA. \\
RoPE / precision & Fixed YaRN factor four, original size 32,768, original
RoPE base. bf16 model/KVzap weights; released mixed bf16/fp32 Fast KVzip gates. \\
KVzap-MLP & \texttt{nvidia/KVzap-mlp-Qwen3-8B}, revision
\texttt{bd5c59178466}. Replace fixed threshold with global fixed-budget
selection; protect the final 128 tokens. \\
Fast KVzip & \texttt{Jang-Hyun/Fast-KVzip}, revision
\texttt{67ce1265f510}, \texttt{q4\_dim16\_sink16.pt} gates.
Global allocation; promote first four and last 4,096 token scores.
KVPress full prefill replaces the released implementation's eviction between
prefill chunks. \\
Repetitions & Five seeded, interleaved repetitions after one warm-up per
method; plain-prefill baseline in each repetition; CUDA synchronized at both
boundaries. \\
Timed region & Per-request setup, scoring, selection, physical compaction.
Exclude loading, tokenization, compilation/warm-up, validation, and generation. \\
Packed cache & Head-major, chronological K/V pairs with token positions and
per-head offsets; untimed checks of counts, finite tensors, ordered positions,
and compactor attention outputs. \\
\bottomrule
\end{tabular}
\end{table}

\section{First-chunk state measurements}
\label{app:first_chunk}

Figure~\ref{fig:first_chunk_mechanism} uses eight LongBench contexts per model
(bf16, SDPA), Section~\ref{sec:method}'s prompts/chunks, and full-cache,
unpruned reconstruction. Query differences are $1-\cos$ after normalization
and before RoPE, averaged over 256 evenly spaced token offsets shared by first/later full
chunks, query heads, and layers 1 onward; layer 0 checks alignment.

The prefix control fixes second-chunk tokens, positions, and reconstruction
reference. Fresh prefill retains the chat prefix, target chunk, and last
$b\in\{0,256,1024,\mathrm{Full}\}$ preceding tokens; Full is 2,032/2,036 tokens
for Qwen/Llama. This isolates context availability; natural chunks also differ
in text and repeat prompts.

Table~\ref{tab:first_chunk_measurements} aggregates finite per-layer/KV-head
Spearman correlations over positions within each chunk. Section~\ref{sec:analysis}
instead correlates entire-context scores and takes the median across layers
and heads. These diagnostics motivate reconstructing the region with greatest
query mismatch, without establishing an optimal calibration chunk.

\begin{table}[H]
\centering
\caption{First-chunk diagnostics, equally weighting contexts after averaging later
full chunks within each context; partial final chunks are excluded. Arrows
compare first with later chunks, except the prefix control ($0\to$ Full).
The median row replaces only the within-chunk mean over layers and KV heads.}
\label{tab:first_chunk_measurements}
\small
\setlength{\tabcolsep}{5pt}
\begin{tabular}{@{}lll@{}}
\toprule
Measurement & Qwen3-8B & Llama-3.1-8B \\
\midrule
Query difference, first $\to$ later
& $\FcQueryFirstQwen{}\to\FcQueryLaterQwen{}$
& $\FcQueryFirstLlama{}\to\FcQueryLaterLlama{}$ \\
Query difference, prefix $0\to$ Full
& $\FcPrefixZeroQwen{}\to\FcPrefixFullQwen{}$
& $\FcPrefixZeroLlama{}\to\FcPrefixFullLlama{}$ \\
Mean within-chunk Spearman $\rho$
& $\FcRhoFirstQwen{}\to\FcRhoLaterQwen{}$
& $\FcRhoFirstLlama{}\to\FcRhoLaterLlama{}$ \\
Median within-chunk Spearman $\rho$
& $\FcMedianRhoFirstQwen{}\to\FcMedianRhoLaterQwen{}$
& $\FcMedianRhoFirstLlama{}\to\FcMedianRhoLaterLlama{}$ \\
\bottomrule
\end{tabular}
\end{table}

\end{document}